\documentclass[11pt]{article}

\usepackage[utf8]{inputenc}
\usepackage[T1]{fontenc}
\usepackage{hyperref}
\usepackage{url}
\usepackage{booktabs}
\usepackage{amsfonts}
\usepackage{nicefrac}
\usepackage{microtype}

\usepackage{cite}
\usepackage{amsmath,amssymb}
\usepackage{graphicx}
\usepackage{dcolumn}
\usepackage{multirow}
\usepackage{colortbl}
\usepackage{siunitx}
\graphicspath{ {./images/}, {./figures/} }
\usepackage{float}
\usepackage{caption}
\usepackage{subcaption}
\usepackage{textcomp}
\usepackage{array}
\usepackage{xcolor}
\usepackage{algorithm}
\usepackage{algpseudocode}
\usepackage{appendix}

\usepackage[margin=1in]{geometry}

\usepackage{authblk}

\title{Deep Learning--Based Fixation Type Prediction for Quality Assurance in Digital Pathology}

\author[1,2,3]{Oskar Thaeter}
\author[4]{Tanja Niedermair}
\author[5]{Jan E.G. Albin}
\author[6,7]{Johannes Raffler}
\author[8,9]{Ralf Huss}
\author[1,2,3,10]{Peter J. Sch\"uffler}

\affil[1]{Institute of Pathology, Technical University of Munich, Germany}
\affil[2]{School of Computation, Information and Technology, Technical University of Munich, Germany}
\affil[3]{Munich Center for Machine Learning, Germany}
\affil[4]{Institute of Pathology, University of Regensburg, Regensburg, Germany}
\affil[5]{Institute of Pathology, Medical Faculty Mannheim, University Heidelberg, Germany}
\affil[6]{Digital Medicine, University Hospital of Augsburg, Germany}
\affil[7]{Bavarian Cancer Research Center (BZKF), Augsburg, Germany}
\affil[8]{Institute of Pathology and Molecular Diagnostics, University Hospital of Augsburg, Germany}
\affil[9]{Current address: Bio$^\text{M}$ Biotech Cluster Development GmbH, Martinsried, Germany}
\affil[10]{Munich Data Science Institute, Technical University of Munich, Germany}

\date{}

\begin{document}

\maketitle

\begin{abstract}
Accurate annotation of fixation type is a critical step in slide preparation for pathology laboratories. 
However, this manual process is prone to errors, impacting downstream analyses and diagnostic accuracy. 
Existing methods for verifying formalin-fixed, paraffin-embedded (FFPE), 
and frozen section (FS) fixation types typically require full-resolution whole-slide images (WSIs), 
limiting scalability for high-throughput quality control.

We propose a deep-learning model to predict fixation types using low-resolution, pre-scan thumbnail images. 
The model was trained on WSIs from the TUM Institute of Pathology (n=1,200, Leica GT450DX) 
and evaluated on a class-balanced subset of The Cancer Genome Atlas dataset (TCGA, n=8,800, Leica AT2), 
as well as on class-balanced datasets from Augsburg (n=695 [392 FFPE, 303 FS], Philips UFS) 
and Regensburg (n=202, 3DHISTECH P1000).

Our model achieves an AUROC of $0.88$ on TCGA, outperforming comparable pre-scan methods by $4.8\%$. 
It also achieves AUROCs of $0.72$ on Regensburg and Augsburg slides, pointing to scanner-induced domain shift as an open challenge. 
Furthermore, the model processes each slide in $21$ ms, $400\times$ faster than existing high-magnification, 
full-resolution methods, enabling rapid, high-throughput processing.

This approach provides an efficient solution for detecting labelling errors without relying on high-magnification scans, 
offering a valuable tool for quality control in high-throughput pathology workflows. 
Future work will improve and evaluate the model's generalisation to additional scanner types. 
Our findings suggest that this method can increase accuracy and efficiency in digital pathology workflows 
and may be extended to other low-resolution slide annotations.

\end{abstract}

\section{Introduction}\label{section:introduction}

\begin{figure}[t]
    \centering

    \begin{subfigure}[t]{0.65\textwidth}
        \centering
        \caption{\textbf{FFPE}}
        \includegraphics[width=1.0\textwidth, keepaspectratio]{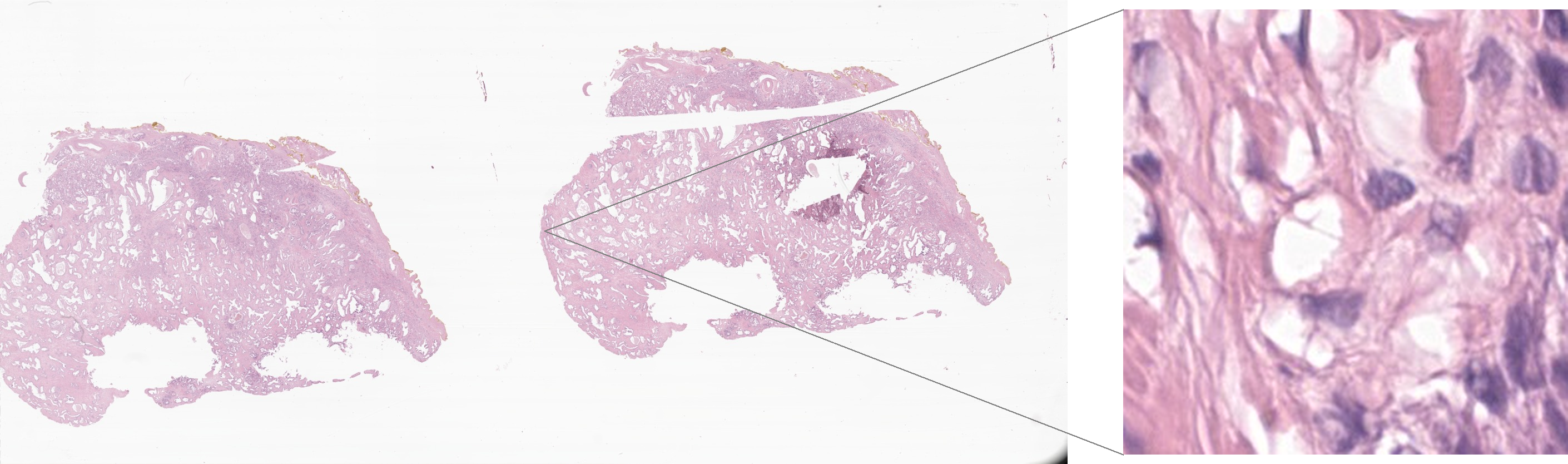}
    \end{subfigure}

    \begin{subfigure}[t]{0.65\textwidth}
        \centering
        \caption{\textbf{FS}}
        \includegraphics[width=1.0\textwidth, keepaspectratio]{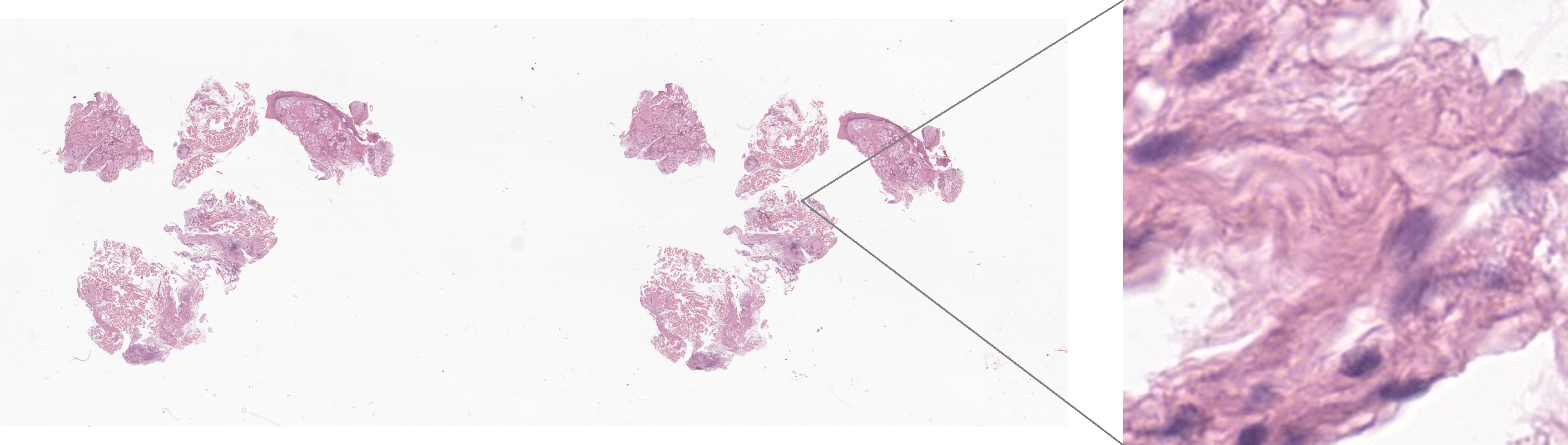}
    \end{subfigure}
    \caption{Example slides of FFPE and FS fixation types shown as thumbnails and at high magnification.}
    \label{fig:fixationtype_examples}
\end{figure}

\begin{figure}[h]
    \centering
    \begin{subfigure}[t]{0.85\textwidth}
        \centering
        \includegraphics[width=\textwidth, keepaspectratio]{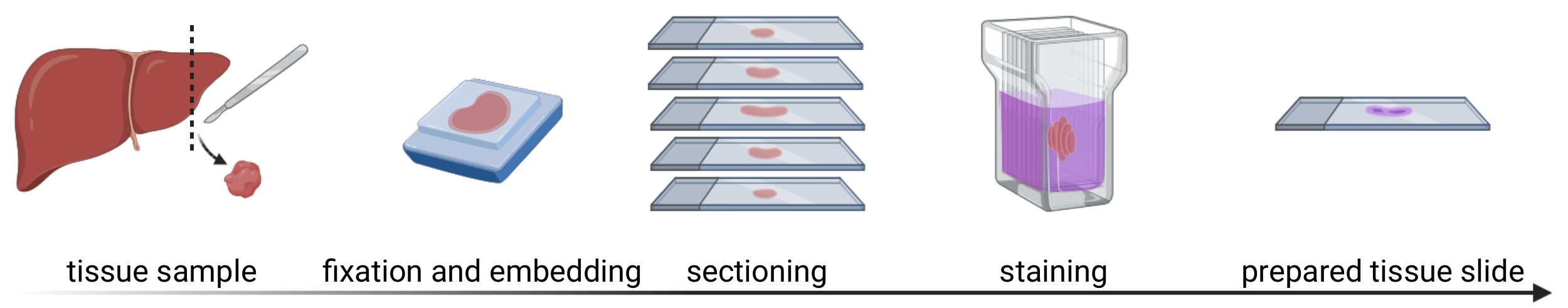}
        \caption{Overview of the tissue processing workflow.}
        \label{fig:tissue_processing_overview}
    \end{subfigure}

    \vspace{1em}

    \begin{subfigure}[t]{0.48\textwidth}
        \centering
        \includegraphics[width=\textwidth, keepaspectratio]{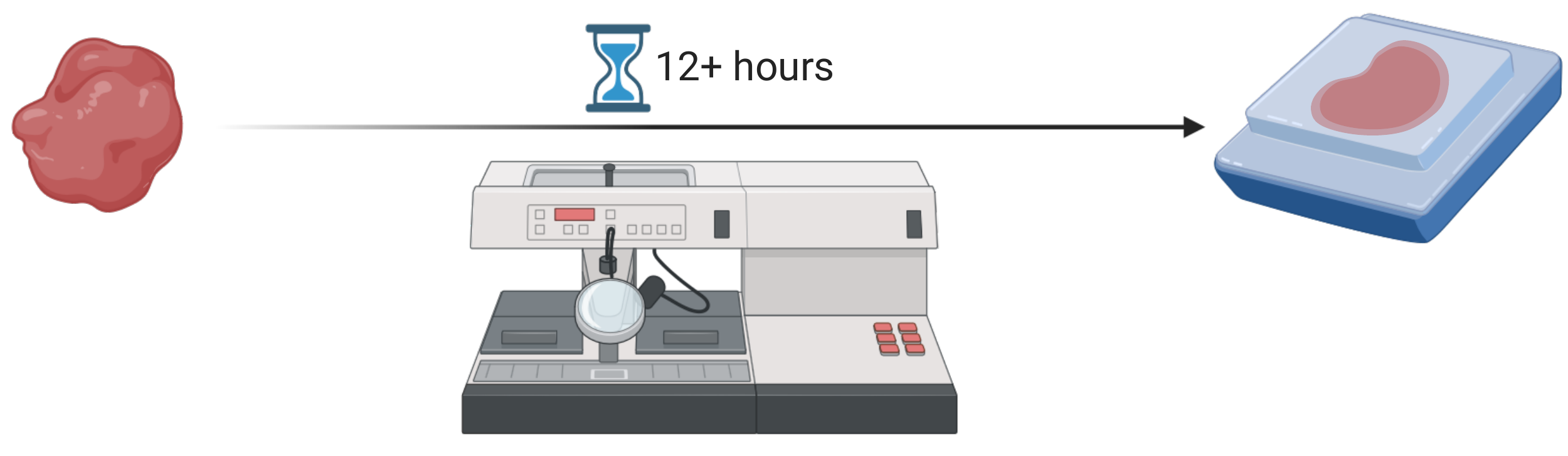}
        \caption{Formalin-fixed, paraffin-embedded (FFPE) process.}
        \label{fig:FFPE_fixation}
    \end{subfigure}
    \hfill
    \begin{subfigure}[t]{0.48\textwidth}
        \centering
        \includegraphics[width=\textwidth, keepaspectratio]{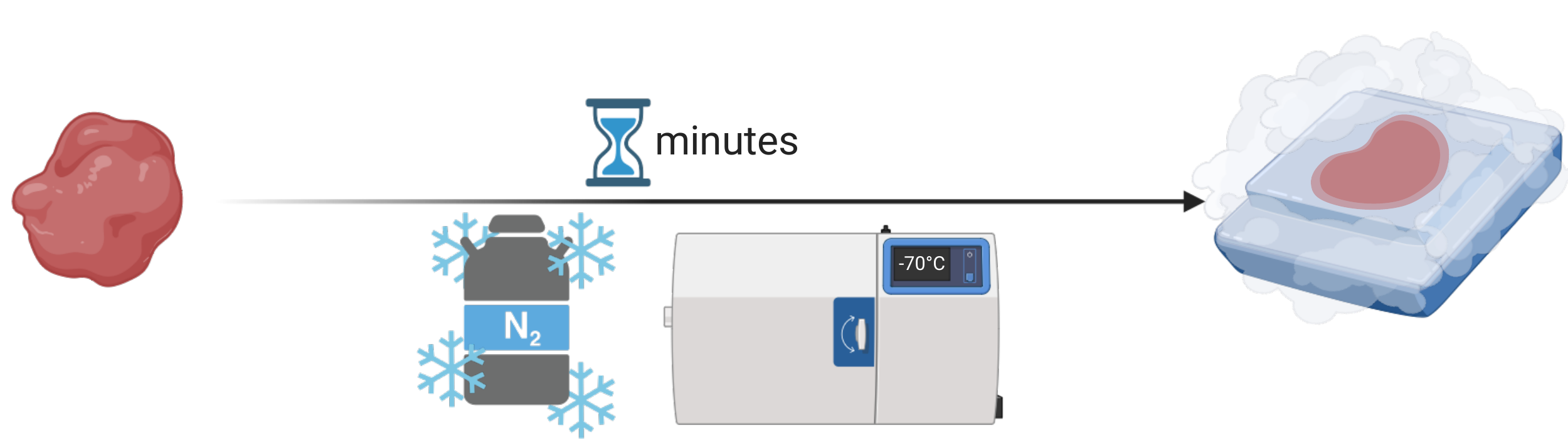}
        \caption{Frozen section (FS) process.}
        \label{fig:FS_fixation}
    \end{subfigure}
    \caption{Tissue processing workflow and the two fixation methods used in this study.
        (a) Tissue samples are fixed, sectioned, mounted, and stained.
        (b) In FFPE processing, tissue is formalin-fixed and paraffin-embedded before microtome sectioning.
        (c) In FS processing, tissue is snap-frozen and sectioned in a cryostat.
        \\ \footnotesize{\textit{Created with BioRender.com}}}
    \label{fig:tissue_processing}
\end{figure}

Digital pathology has transformed histopathological diagnostics by digitising glass slides into whole-slide images (WSIs), allowing computational analysis of tissue samples \cite{PANTANOWITZ201136}. A critical step in slide preparation is the fixation of tissue, which preserves cellular structures for downstream examination. The two predominant fixation methods are formalin-fixed, paraffin-embedded (FFPE) processing, in which tissue is chemically fixed and embedded in paraffin wax \cite{bussolati2022fixation, fox1985formaldehyde}, and frozen section (FS) processing, in which tissue is snap-frozen for rapid intraoperative assessment \cite{wallace2024} (Figure~\ref{fig:tissue_processing}). Each method produces distinct histological characteristics: FFPE sections exhibit superior morphological preservation, while frozen sections, though faster to prepare, often contain freezing artefacts and less well-preserved cellular architecture (Figure~\ref{fig:fixationtype_examples}).

Accurate fixation-type annotation is essential for reliable histopathological analysis. Jang et al.\ \cite{Jang2021} showed that the performance of deep learning models for cancer diagnosis depends on the fixation type of the training data, which highlights the importance of correct annotations. In practice, mislabelling can occur when handling large volumes of slides, and retrospective correction across extensive physical or digital archives is costly and time-consuming.

Existing computational approaches to fixation-type classification operate on high-resolution WSIs. Dan et al.\ \cite{Dan2022} achieved $98.91\%$ accuracy using a VGG-19 model with majority voting over thousands of tiles extracted at $40\times$ magnification, but this requires at least $10$ seconds per slide. Such methods are impractical for high-throughput quality control or pre-scan verification, where the full-resolution image may not yet be available.

Most WSI file formats store a low-resolution thumbnail image, typically captured before the high-resolution scan and up to $128\times$ smaller than the full image \cite{GOODE201327}. These thumbnails load orders of magnitude faster than the tiled pyramid of the full scan. If fixation type could be reliably predicted from these thumbnails alone, mislabelled slides could be flagged, re-annotated, or excluded before the high-resolution scan is even conducted, enabling efficient pre-scan quality control. Predicting fixation type at low magnification is difficult, however, because the cellular-level features that distinguish FFPE from FS sections are not resolved at thumbnail scale.

To our knowledge, the only prior work using thumbnail-level images for fixation-type prediction is that of Weng et al.\ \cite{Weng2019MultimodalMR}, who employed a multi-modal, multi-task learning approach incorporating slide thumbnails, tile images, free text, and structured metadata. Their thumbnail-only model achieved an AUROC of $0.78$, and their best model without high-magnification tiles reached $0.84$, suggesting that fixation type is difficult to predict at the slide level alone.

\begin{figure}[t]
    \centering
    \includegraphics[width=1.0\textwidth, keepaspectratio]{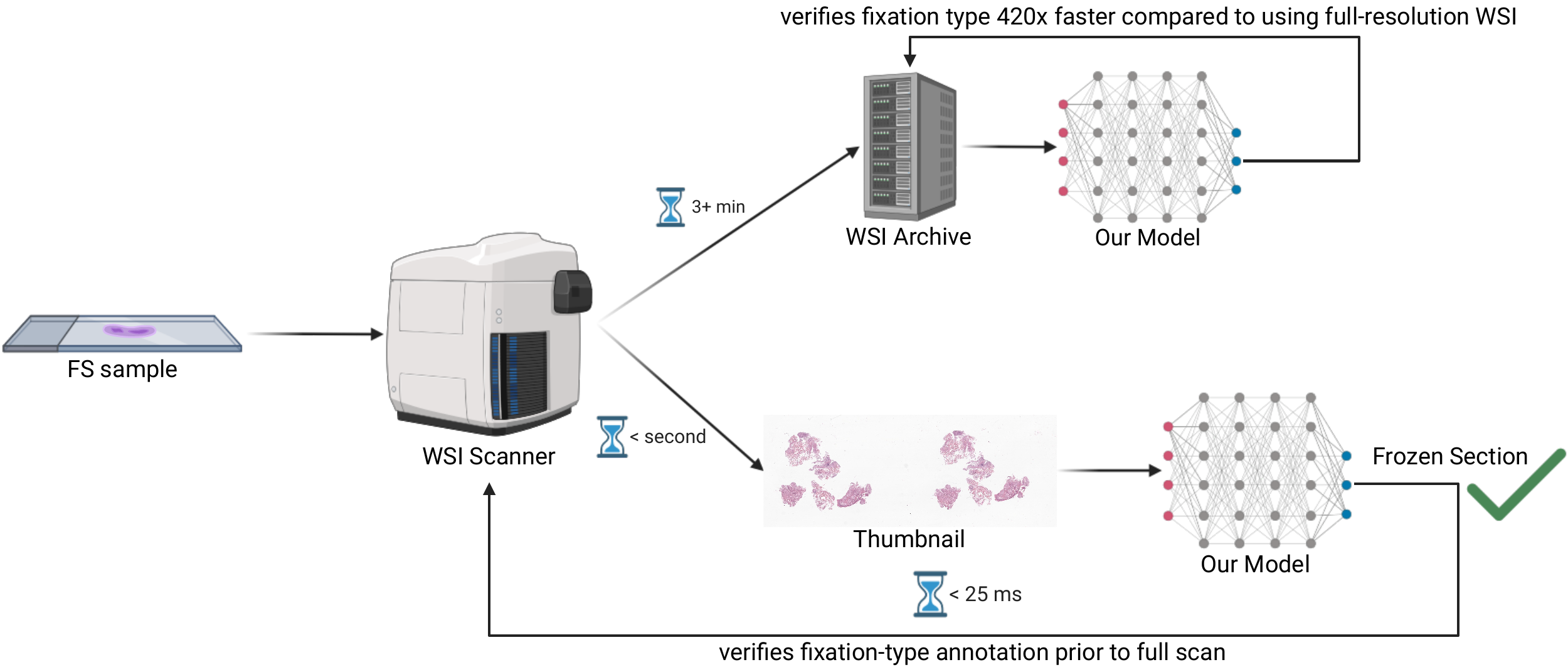}
    \caption{Possible applications of our approach in a digital pathology workflow.
        The fixation-type annotation can be verified before the high-resolution scan is conducted,
        allowing for re-annotation, re-examination, or skipping the scan entirely.
        Our approach can also be used as an efficient quality control measure in digital pathology archives.
        \\ \footnotesize{\textit{Created with BioRender.com}}}
    \label{fig:OurApproach}
\end{figure}

In this work, we propose a deep learning model that predicts fixation type using only low-resolution pre-scan thumbnail images. We evaluate four pathology-pretrained vision transformer backbones and compare whole-slide and tiled-slide classification strategies. Trained on data from the TUM Institute of Pathology ($n=1{,}200$, Leica GT450DX), our best model achieves an AUROC of $0.88$ on a class-balanced subset of The Cancer Genome Atlas (TCGA, $n=8{,}800$, Leica AT2), improving upon Weng et al.'s thumbnail-based result by $12.8\%$. The model processes each slide in $21$\,ms, which is $400\times$ faster than high-magnification methods, making it suitable for high-throughput pre-scan quality control. We further evaluate cross-scanner generalisation on datasets from Augsburg (Philips UFS) and Regensburg (3DHISTECH P1000), identifying scanner-induced domain shift as a key challenge for future work. Figure~\ref{fig:OurApproach} illustrates the envisioned application of our approach.

\section{Related Work}\label{section:related work}

Computational quality control in digital pathology has primarily targeted scan-level artefacts rather than annotation correctness. Campanella et al.\ \cite{Campanella2017TowardsML} developed a high-throughput blur detection system, and Haghighat et al.\ \cite{Haghighat2021PathProfilerAQ} proposed an automatic quality assessment pipeline producing focus and H\&E staining quality scores for large-scale histology cohorts. These methods address image quality but do not verify slide-level metadata such as fixation type.

Fixation-type classification has received comparatively little attention. Dan et al.\ \cite{Dan2022} fine-tuned a VGG-19 model on thousands of tiles per slide at $40\times$ magnification, aggregating predictions via majority voting to achieve $98.91\%$ accuracy. While highly accurate, this approach requires the full-resolution scan and takes at least $10$ seconds per slide, precluding pre-scan use.

Weng et al.\ \cite{Weng2019MultimodalMR} addressed slide-level metadata prediction, including fixation type, through a multi-modal, multi-task framework combining a $0.3125\times$ slide thumbnail ($512 \times 512$\,px), three tile images at $5\times$ magnification ($299 \times 299$\,px), free text, and structured case-level data. Using ResNet-50 for image representation, they found that fixation type was difficult to predict from the slide thumbnail alone (AUROC $0.78$) but considerably easier with high-magnification tiles, attributing this to the loss of intracellular matrix detail in frozen sections at low resolution. Their work represents the only prior attempt at thumbnail-level fixation-type prediction and serves as our primary benchmark (Section~\ref{subsection:discussion sota}).

\section{Methods}\label{section:methods}

\subsection{Datasets}\label{subsection:datasets}
We used four datasets spanning different scanners and institutions (Table~\ref{tab:datasplits}). The primary training dataset was obtained from the Institute of Pathology at the Technical University of Munich (TUM), comprising $2{,}160$ class-balanced WSIs scanned on a Leica GT450DX. Three external datasets served as held-out test sets: a class-balanced subset of The Cancer Genome Atlas (TCGA, $n=8{,}800$, Leica AT2), slides from Augsburg ($n=695$, Philips UFS), and slides from Regensburg ($n=202$, 3DHISTECH P1000). A portion of the TCGA data was discarded due to missing or corrupted images and to balance the classes.

\begin{table}[tbp]
    \caption{Data splits of each dataset (FFPE, FS).}
    \label{tab:datasplits}
    \centering
    \begin{tabular}{l l l l l}
        \toprule
        Dataset          & Total      & Train       & Validation    & Test          \\
        \midrule
        TUM              & 1080, 1080 & 600, 600    & 240, 240      &  240,  240    \\
        TCGA             & 4400, 4400 & \textemdash & \textemdash   & 4400, 4400    \\
        Augsburg         &  392,  303 & \textemdash & \textemdash   &  392,  303    \\
        Regensburg       &  101,  101 & \textemdash & \textemdash   &  101,  101    \\
        \bottomrule
    \end{tabular}
\end{table}

FFPE thumbnail examples from each external dataset, alongside magnified regions at full resolution, are shown in Figure~\ref{fig:dataset_examples} (Appendix).

\subsection{Pre-Processing}\label{subsection:pre-processing}

\begin{figure}[H]
    \centering
    \includegraphics[width=0.8\textwidth]{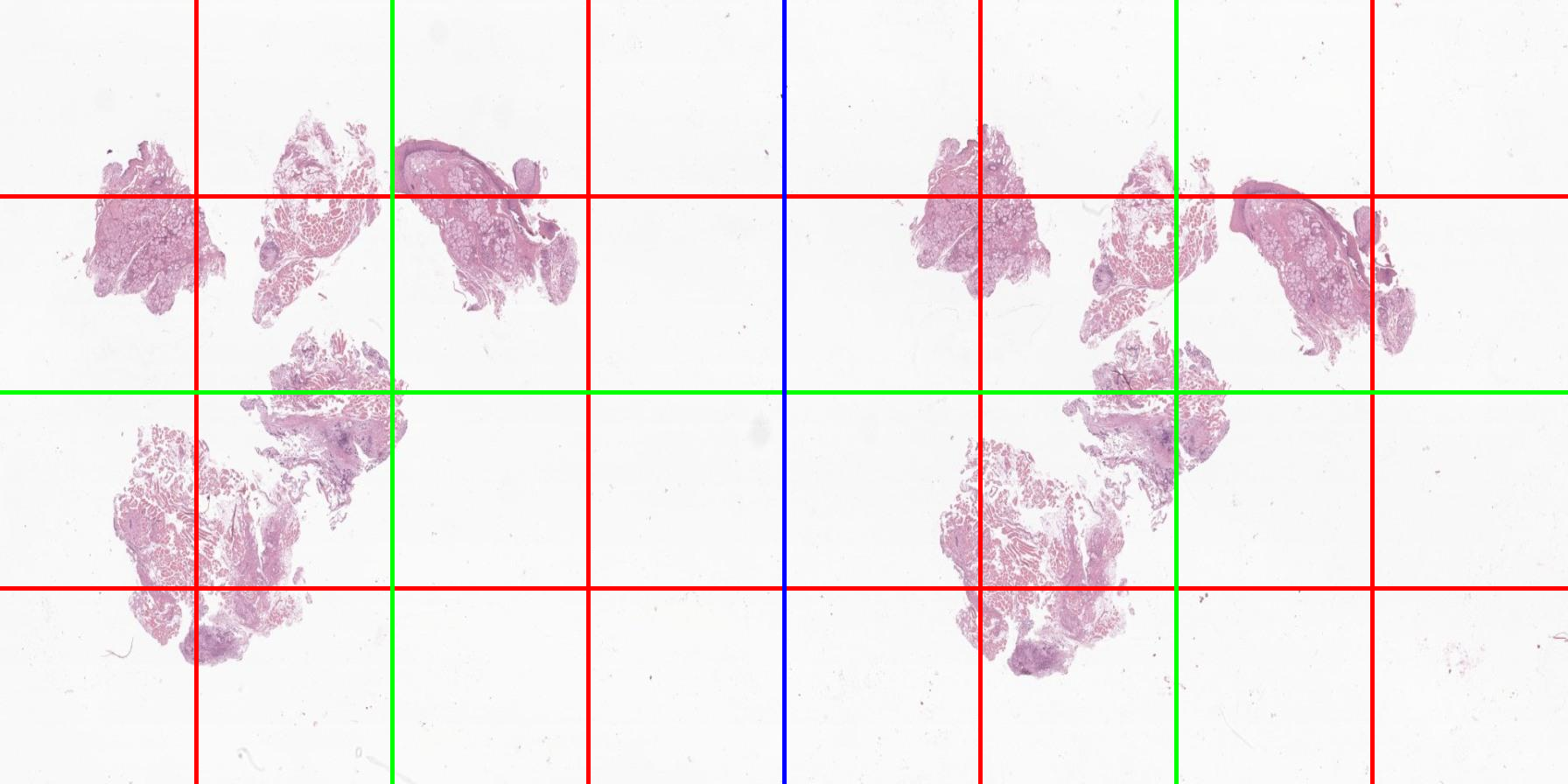}
    \caption{Stretched slide thumbnail with tile grids overlaid. Red: L grid ($4 \times 8$), green: M grid ($2 \times 4$), blue: S grid ($1 \times 2$).}
    \label{fig:tilegrid}
\end{figure}

\begin{figure}[H]
    \centering
    \includegraphics[width=1.0\textwidth]{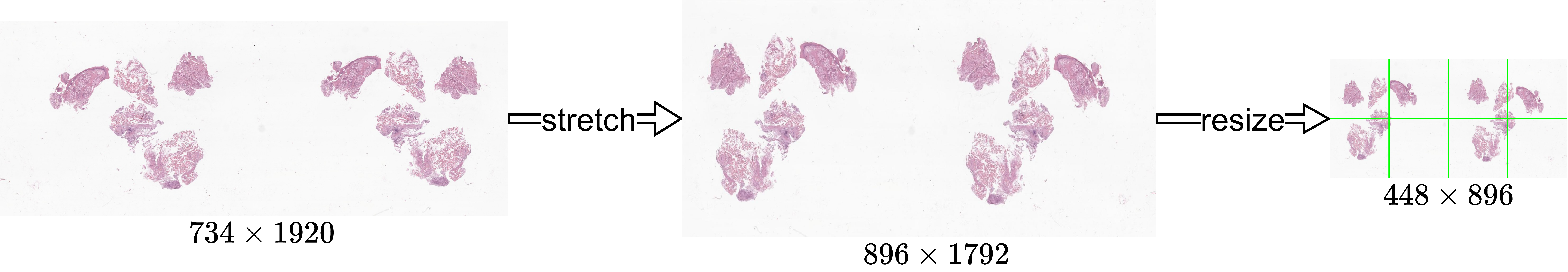}
    \caption{Pre-processing workflow for the M configuration. The thumbnail is stretched to $896 \times 1{,}792$\,px, resized to $448 \times 896$\,px, and divided into a $2 \times 4$ grid of $224 \times 224$\,px tiles (green).}
    \label{fig:preprocessing_workflow}
\end{figure}

Using the OpenSlide library \cite{GOODE201327}, we extracted a thumbnail from each WSI: either the auxiliary thumbnail image stored in the file or, if unavailable, the lowest pyramid level resized so that its longest side is $1{,}920$\,px. All thumbnails were oriented so that the width exceeded the height and stretched to a common size of $896 \times 1{,}792$\,px. Depending on the desired resolution, the images were then resized to one of four scales (Table~\ref{tab:slideresolutions}) and divided into a grid of non-overlapping $224 \times 224$\,px tiles (Figure~\ref{fig:tilegrid}). The complete workflow is illustrated in Figure~\ref{fig:preprocessing_workflow} for the M configuration.

\begin{table}[htb]
    \caption{Thumbnail resolutions and corresponding tile grid configurations.}
    \label{tab:slideresolutions}
    \centering
    \begin{tabular}{c c c}
        \toprule
        Name & Height $\times$ Width (px) & Grid (rows $\times$ cols) \\
        \midrule
        XS   & $224 \times 224$           & $1 \times 1$          \\
        S    & $224 \times 448$           & $1 \times 2$          \\
        M    & $448 \times 896$           & $2 \times 4$          \\
        L    & $896 \times 1792$          & $4 \times 8$          \\
        \bottomrule
    \end{tabular}
\end{table}

\subsection{Vision Backbones}\label{subsection:visionbackbone}

We compare four pathology-pretrained vision transformers as feature extractors (Table~\ref{tab:visionbackbone}): TransPath \cite{wang2021transpath} (ViT-S/16, pretrained on PAIP and TCGA via MoCo v3 \cite{chen2021mocov3}), UNI \cite{chen2024uni} (ViT-L/16, trained on 100M private images that do not include TCGA, avoiding data contamination), Virchow2 \cite{Zimmermann2024Virchow2SS} (ViT-H/14, trained with DINOv2 \cite{Oquab2023DINOv2LR} on 3.1M WSIs from Memorial Sloan Kettering), and H-Optimus-0 \cite{hoptimus0} (ViT-g/14, trained with iBOT \cite{Zhou2021iBOTIB}/DINOv2 on over 500K H\&E WSIs, using register tokens \cite{Darcet2023VisionTN}).

\begin{table}[htpb]
    \caption{Overview of the vision transformer backbones evaluated in this study.}
    \label{tab:visionbackbone}
    \centering
    \begin{tabular}{cccccc}
        \toprule
        Name        & Architecture & \#M Parameters           & Embedding           & Output \\
        \midrule
        TransPath   & ViT-S/16     &  \phantom{11}22          & \phantom{1}384      & class token \\
        UNI         & ViT-L/16     &  \phantom{1}303          & 1024                & \multirow{3}{*}{class + mean patch token} \\
        Virchow2    & ViT-H/14     &  \phantom{1}631          & 1280                & \\
        H-Optimus-0 & ViT-g/14     &  1100                    & 1536                & \\
        \bottomrule
    \end{tabular}
\end{table}

\subsection{Classification Approaches}\label{subsection:classification_approaches}

We evaluate two families of approaches for slide-level fixation-type prediction from thumbnail images.

\subsubsection{Whole-Slide Classification}\label{subsubsection:wholeslide}
In the simplest variant, \textbf{``XS Slides''}, the thumbnail is resized to $224 \times 224$\,px and the entire backbone is fine-tuned end-to-end. Alternatively, in \textbf{``ViT Upscaling''}, we interpolate the position embeddings to accommodate the larger token count of an M-sized ($448 \times 896$\,px) thumbnail, preserving pretrained weights while fine-tuning only the attention and position embedding layers \cite{Touvron2022ThreeTE}.

\subsubsection{Tiled-Slide Classification}\label{subsubsection:tiledslide}
This approach tiles the thumbnail into $224 \times 224$\,px patches processed independently by the backbone. We compared M (8 tiles) and L (32 tiles) configurations and found more tiles to be generally beneficial; we therefore used L ($4 \times 8 = 32$ tiles). To obtain a slide-level prediction, we evaluated three aggregation strategies:

\begin{itemize}
    \item \textbf{Soft voting}: averages tile-level sigmoid predictions, preserving differentiability for end-to-end training.
    \item \textbf{Multi-head attention}: computes attention weights $\alpha_i$ over tile-level feature vectors $\mathbf{f}_i \in \mathbb{R}^D$ and returns the weighted sum $\mathbf{F}_{\text{att}} = \sum_{i=1}^{n} \alpha_i \mathbf{f}_i$.
    \item \textbf{Transformer}: treats tile features as tokens with learned position embeddings; the class token is used for classification.
\end{itemize}

\subsection{Classification Head}\label{subsection:classifierhead}

All approaches use a three-layer feed-forward classification head with batch normalisation \cite{Ioffe2015BatchNA}, ReLU activations, and dropout ($p = 0.1$) \cite{Hinton2012ImprovingNN}:
\[
    \texttt{HiddenLayer}(x; W, b) = \texttt{Dropout}(\texttt{ReLU}(\texttt{BatchNorm}(\mathbf{W}x + b)))
\]
The final layer outputs a single logit passed through a sigmoid function. Layer sizes for each backbone (Table~\ref{tab:classifiers}) were selected via Bayesian hyperparameter optimisation using a Tree-structured Parzen Estimator \cite{Bergstra2011AlgorithmsFH} with Hyperband pruning \cite{Li2016HyperbandAN} over 256 trials per backbone.

\begin{table}[htpb]
    \caption{Classification head layer sizes for each backbone, selected via hyperparameter optimisation.}
    \label{tab:classifiers}
    \centering
    \begin{tabular}{l r r r}
        \toprule
        Backbone    & 1st Layer & 2nd Layer & 3rd Layer \\
        \midrule
        TransPath   &   2048    &  1920     &      128  \\
        UNI         &   1600    &    64     &      192  \\
        Virchow2    &   1728    &    64     &      192  \\
        H-Optimus-0 &   1856    &   192     &      128  \\
        \bottomrule
    \end{tabular}
\end{table}

\section{Results}\label{section:results}

\subsection{Model Selection}\label{subsection:model selection}

Table~\ref{tab:eval results} reports the validation accuracy on the TUM dataset for all backbone--approach combinations. Tiled-slide approaches consistently outperformed whole-slide approaches across all backbones, with soft voting yielding the highest average accuracy ($87\%$). Among backbones, UNI achieved the best average accuracy ($87\%$), closely followed by Virchow2 and H-Optimus-0. The best overall configuration was UNI with soft voting, reaching $89\%$ validation accuracy.

\begin{table}[htbp]
    \centering
    \caption{TUM validation accuracy across backbones and classification approaches.}
    \label{tab:eval results}
    \begin{tabular}{lccccc}
        \toprule
                    & \multicolumn{2}{c}{Whole-Slide}   & \multicolumn{3}{c}{Tiled-Slide}                   \\
        \cmidrule(lr){2-3}
        \cmidrule(lr){4-6}
                    & XS Slides & ViT Upscaling         & soft voting   & multiattention    & transformer   \\
        \midrule
        TransPath   & 0.83      & 0.85                  & 0.87          & 0.88              & 0.87          \\
        UNI         & 0.85      & 0.86                  & 0.89          & 0.87              & 0.88          \\
        Virchow2    & 0.86      & 0.88                  & 0.88          & 0.87              & 0.86          \\
        H-Optimus-0 & 0.84      & 0.86                  & 0.88          & 0.87              & 0.85          \\
        \bottomrule
    \end{tabular}
\end{table}

Table~\ref{tab:latency} shows the corresponding single-threaded inference latencies. Latency scales with both backbone size and input resolution: XS Slides is the fastest, ViT Upscaling approximately $7\times$ slower, and tiled-slide approaches approximately $12\times$ slower. The three tiled-slide aggregation strategies add negligible overhead relative to the backbone forward passes. For the selected UNI--soft voting model, inference takes $21$\,ms per slide.

\begin{table}[htbp]
    \centering
    \caption{Single-threaded inference latency (ms) across backbones and approaches.}
    \label{tab:latency}
    \begin{tabular}{lccccc}
        \toprule
                    & \multicolumn{2}{c}{Whole-Slide}   & \multicolumn{3}{c}{Tiled-Slide}                   \\
        \cmidrule(lr){2-3}
        \cmidrule(lr){4-6}
                    & XS Slides & ViT Upscaling         & soft voting   & multiattention    & transformer   \\
        \midrule
        TransPath   & 0.88      &  4.15                 &  4.26         &  4.31             &  4.67         \\
        UNI         & 2.37      & 15.23                 & 20.87         & 20.95             & 21.85         \\
        Virchow2    & 4.28      & 34.91                 & 56.49         & 56.63             & 57.74         \\
        H-Optimus-0 & 6.64      & 48.54                 & 82.13         & 82.26             & 83.66         \\
        \bottomrule
    \end{tabular}
\end{table}

\subsection{External Evaluation}\label{subsection:test results}

We evaluated the selected UNI--soft voting model on the held-out TUM test set and the three external datasets (Table~\ref{tab:test results}). On TUM, the model achieved an accuracy of $89\%$ and an AUROC of $0.94$. Performance generalised well to the TCGA dataset (accuracy $81\%$, AUROC $0.88$), which was scanned on a different Leica model. Performance dropped on slides from non-Leica scanners, however: accuracy fell to $56\%$ on Augsburg (Philips) and $50\%$ on Regensburg (3DHISTECH), with AUROCs of $0.72$ for both, indicating a strong scanner-dependent domain shift.

\begin{table}[htbp]
    \centering
    \caption{Test results for the UNI--soft voting model on internal and external datasets.}
    \label{tab:test results}
    \begin{tabular}{ll|ccc}
        \toprule
        Dataset     & Scanner           & Acc   & F1    & AUROC \\
        \midrule
        TUM         & Leica GT 450 DX   & 0.89  & 0.89  & 0.94  \\
        TCGA        & Leica AT2         & 0.81  & 0.83  & 0.88  \\
        Augsburg    & Philips UFS       & 0.56  & 0.65  & 0.72  \\
        Regensburg  & 3DHISTECH P1000   & 0.50  & 0.66  & 0.72  \\
        \bottomrule
    \end{tabular}
\end{table}

\section{Discussion}\label{section:discussion}

\subsection{Comparison to Prior Work}\label{subsection:discussion sota}

Our thumbnail-only model achieves an AUROC of $0.88$ on TCGA, outperforming Weng et al.'s \cite{Weng2019MultimodalMR} single-task thumbnail-based model ($0.78$, $+12.8\%$) and their best model without high-magnification tiles ($0.84$, $+4.8\%$). It approaches their full multi-modal result ($0.91$), which relies on high-magnification patch images unavailable before scanning. Compared to Dan et al.'s \cite{Dan2022} full-resolution approach ($98.91\%$ accuracy at $40\times$), our method trades $10$ percentage points of accuracy for a $400\times$ reduction in inference time ($21$\,ms vs.\ ${\sim}10$\,s per slide), making it practical for pre-scan quality control where speed is essential.

These results confirm that pre-scan fixation-type classification from thumbnails is feasible and that the accuracy--efficiency tradeoff strongly favours low-resolution approaches for high-throughput workflows.

\subsection{Limitations and Future Work}\label{subsection:limitations}

The main limitation of our approach is scanner-dependent domain shift. While the model generalises well across Leica scanners (TUM to TCGA), performance degrades on slides from Philips and 3DHISTECH devices (AUROC $0.72$). This likely reflects differences in colour calibration, optics, and thumbnail generation across scanner manufacturers. Domain adaptation techniques, such as stain normalisation, scanner-aware training, or multi-site fine-tuning, could address this gap and should be explored in future work.

Beyond scanner generalisation, several directions merit investigation: incorporating additional pre-scan modalities (e.g., barcode metadata or macro images) to complement the thumbnail signal, extending the approach to other slide-level annotations such as staining method, and integrating explainability methods to support adoption in clinical practice.

\section{Conclusion}\label{section:conclusion}

We have presented a deep learning model for fixation-type classification that operates solely on low-resolution pre-scan thumbnail images. On the TCGA dataset, our approach achieves an AUROC of $0.88$, improving upon the prior thumbnail-based state-of-the-art by $12.8\%$, while processing each slide in $21$\,ms. The primary remaining challenge is cross-scanner generalisation, which we have characterised through evaluation on three external datasets from different scanner manufacturers. Our results demonstrate that pre-scan thumbnail analysis is a viable and efficient quality control mechanism for high-throughput digital pathology and can serve as a starting point for broader pre-scan slide annotation verification.

\bibliographystyle{unsrt}
\bibliography{bibliography}

\appendix

\section{Dataset Examples}\label{appendix:dataset_examples}

\begin{figure}[H]
    \centering

    \begin{subfigure}[t]{0.65\textwidth}
        \centering
        \caption{TCGA (Leica Aperio)}
        \includegraphics[width=1.0\textwidth, keepaspectratio]{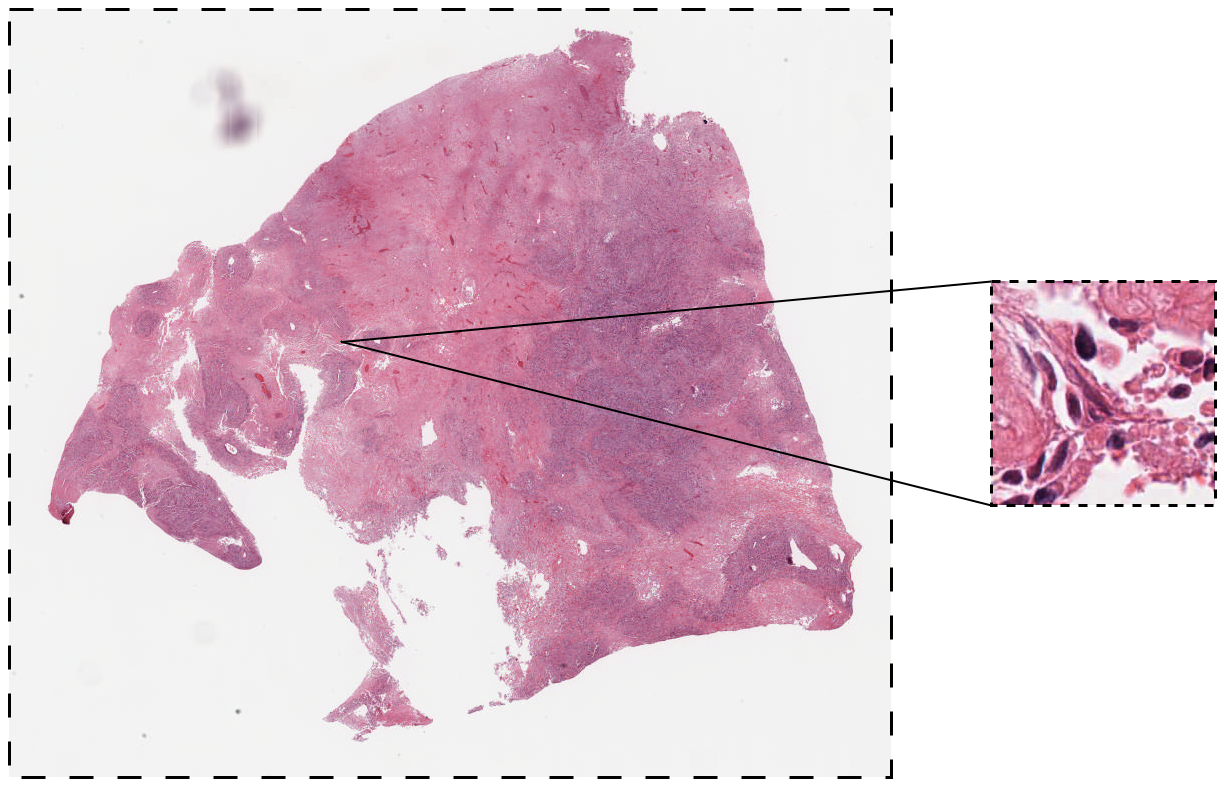}
    \end{subfigure}

    \begin{subfigure}[t]{0.65\textwidth}
        \centering
        \caption{Regensburg (3DHISTECH)}
        \includegraphics[width=1.0\textwidth, keepaspectratio]{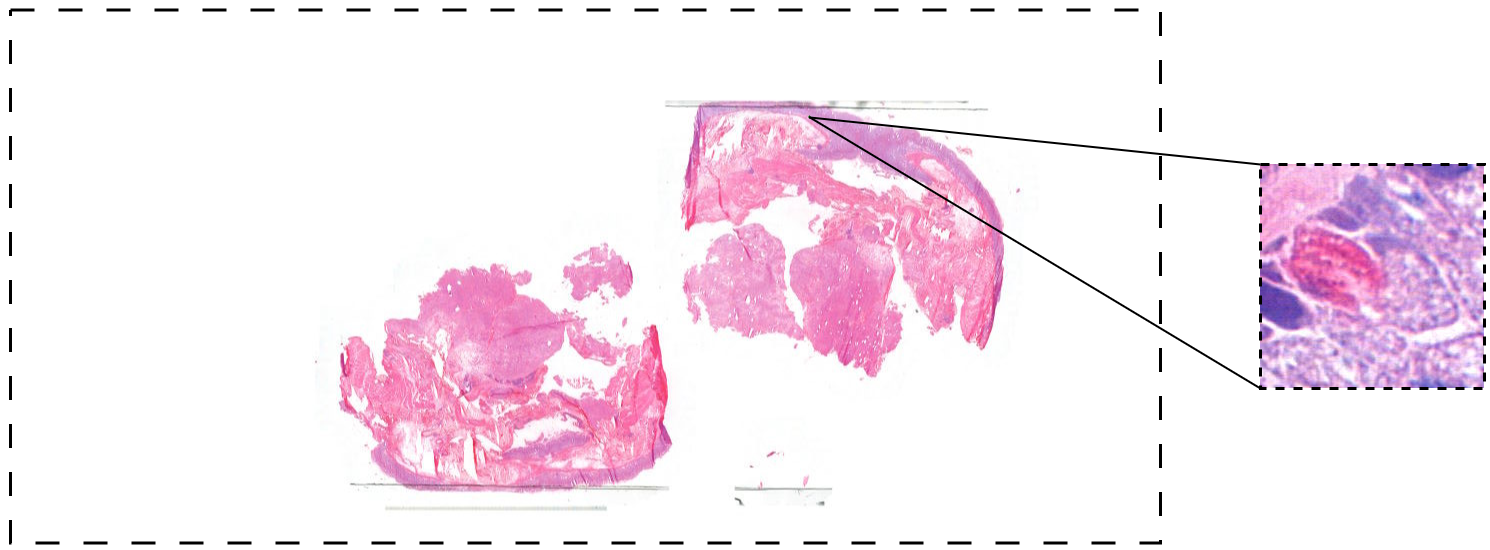}
    \end{subfigure}

    \begin{subfigure}[t]{0.65\textwidth}
        \centering
        \caption{Augsburg (Philips)}
        \includegraphics[width=1.0\textwidth, keepaspectratio]{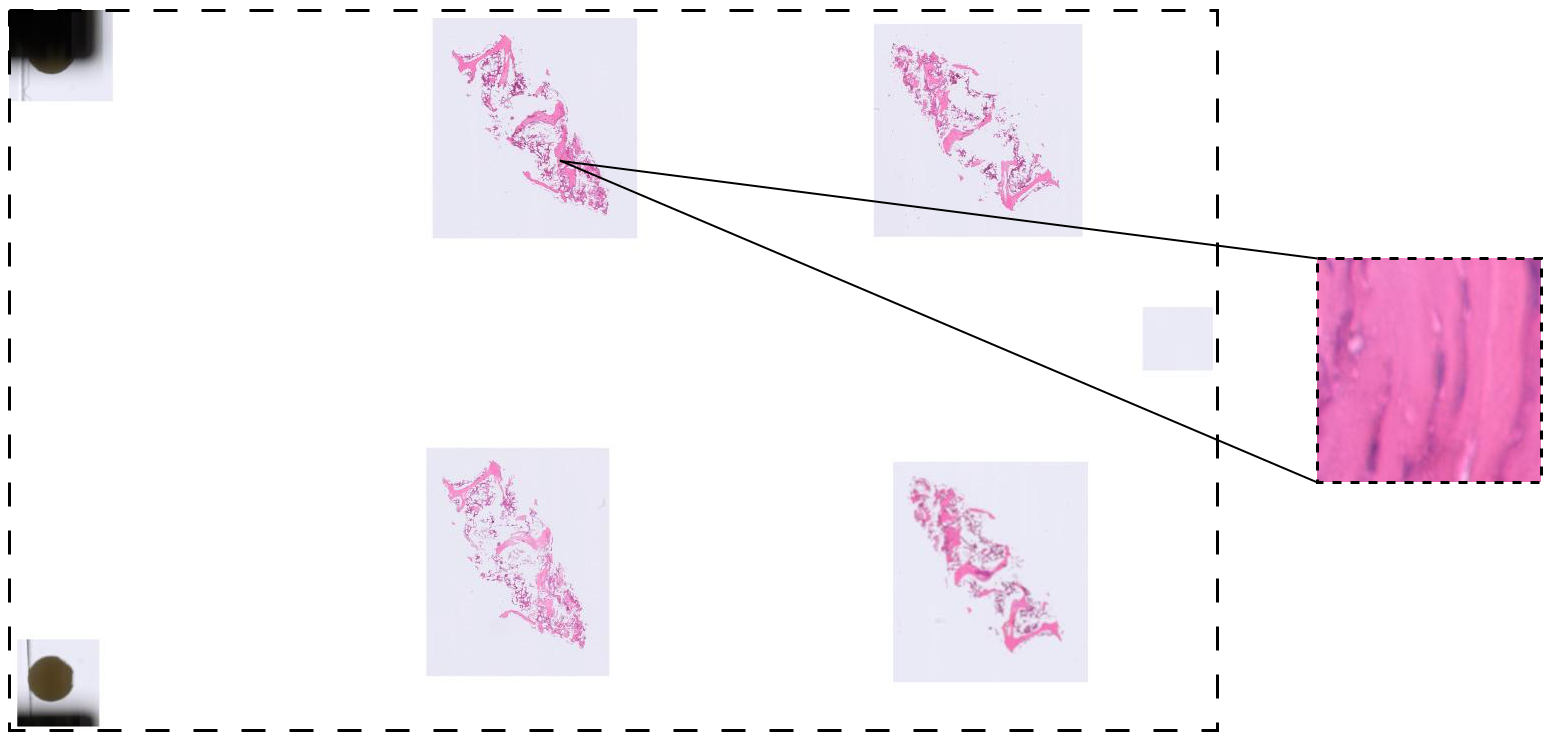}
    \end{subfigure}
    \caption{Thumbnail examples from different scanners with magnified $224 \times 224$\,px regions at full resolution.}
    \label{fig:dataset_examples}
\end{figure}

\end{document}